\documentclass[conference]{IEEEtran}
\IEEEoverridecommandlockouts
\usepackage{cite}
\usepackage{pgfplots}
\usepackage{amsmath,amssymb,amsfonts}
\usepackage{algorithmic}
\usepackage{graphicx}
\usepackage{textcomp}
\usepackage{layouts}
\usepackage{xcolor}
\def\BibTeX{{\rm B\kern-.05em{\sc i\kern-.025em b}\kern-.08em
    T\kern-.1667em\lower.7ex\hbox{E}\kern-.125emX}}
\usepackage{pdfpages}

\begin{document}

\title{Disaster Tweets Classification using BERT-Based Language Model\\
}

\author{\IEEEauthorblockN{Anh Duc Le}
\IEEEauthorblockA{\textit{International School} \\
\textit{Vietnam National University, Hanoi}\\
Building C, HACINCO Student Village, 79 Nguy Nhu Kon Tum, Thanh Xuan, Hanoi, Vietnam \\
ladcva@yahoo.com}
}

\maketitle

\begin{abstract}
Social networking services have became an important communication channel in time of emergency. The aim of this study is to create a machine learning language model that is able to investigate if a person or area was in danger or not. The ubiquitousness of smartphones enables people to announce an emergency they’re observing in real-time. Because of this, more agencies are interested in programatically monitoring Twitter (i.e. disaster relief organizations and news agencies). Design a language model that is able to understand and acknowledge when a disaster is happening based on the social network posts will become more and more necessary over time.
\end{abstract}


\section{Introduction}

\subsection{Overview}
The exponential growth of the Internet, social media, and online platforms has enabled people all over the world to communicate instantly. Every second, on average, around 6,000 tweets are tweeted on Twitter, which corresponds to over 350,000 tweets sent per minute, 500 million tweets per day and around 200 billion tweets per year \cite{b1}. This communication channel has been a concern for agencies since new information is constantly being updated. By combining the rapid update pace from social network and the commodities development of Machine Learning technologies, specifically in Natural Language Processing (NLP), determining when an individual is in danger using posts from Twitter is a binary classification task. Therefore, building a machine learning model to solve this problem by using input text, and provide the result in binary form (Disaster or Not Disaster) automatically can be the effective method to acknowledge and take action to the society.

\subsection{Related works}
The primary objective of this research is to identify the people who in need of assistance based on their social network posts. Recent developments in the field of NLP have taken effect for this specific type of problem with the followings common approaches:
\subsubsection{Traditional approach}
There have been several works proposed for text classification on this labeled dataset with the well-established Logistic Regression (LR) \cite{b2}, Support Vector Machine (SVM) \cite{b3}, Naive Bayes (NB) \cite{b4}, Gradient Boosting (XGB) \cite{b5} and other fundamental machine learning models. However, these methods require manual feature engineering process to encode text sequences in vector form, therefore can be fed into classifiers, which might suffer from the problem of typing error and generates noisy words to the corpus.

\subsubsection{Deep learning approach}
\begin{figure}
  \centering
  \includegraphics[scale=0.5]{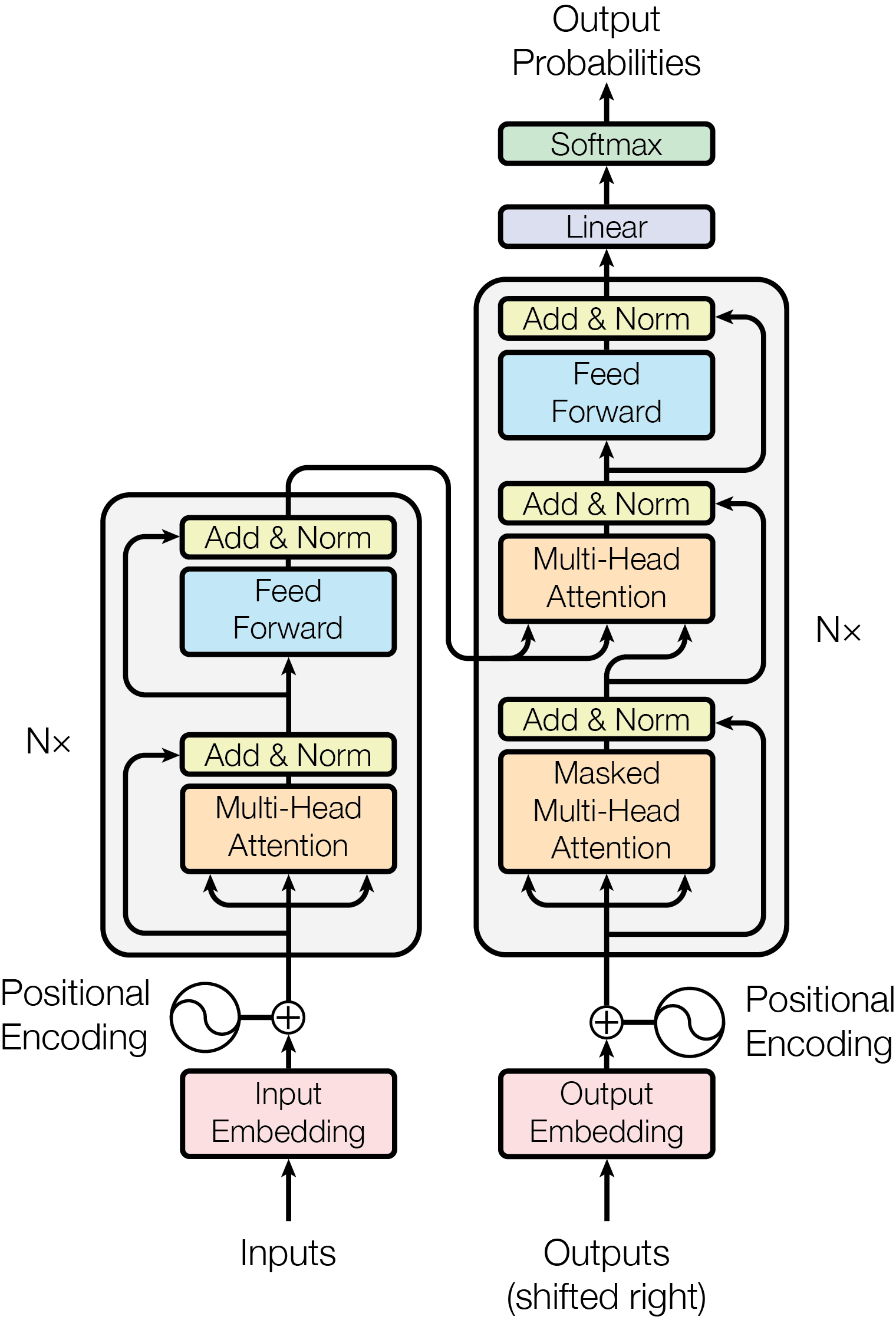}
  \caption{The Transformer - model architecture.}
  \label{fig:model-arch}
\end{figure}

For this approach, researchers have repeatedly shown the value of deep learning for NLP over time. There have been recent developments in the field of NLP, Machine Translation and most of the State Of The Art (SOTA) results have been achieved using Attention models. Topmost models on the GLUE Benchmark Leader board use self-attention models Transformers.

The Transformer neural network architecture proposed by Vaswani et al. \cite{b6} marked one of the major breakthroughs of the decade in the NLP field. The multi-head self-attention layer in Transformer aligns words in a sequence with other words in the sequence, thereby calculating a representation of the sequence. It is not only more effective in representation, but also more computationally efficient compared to convolution and recursive operations.

\section{Solution methods}
\subsection{Data Preprocessing Methods}
There are some mechanism we will manually attempt to our experiment:

\subsubsection{Count Vector} 
is a Frequency-based Embedding, each input document will be one-hot encoded to an $N-dimension$ vector containing count for each word's occurrence, where $N$ is the size of vocabulary.

\subsubsection{Term Frequency – Inverse Document Frequency (TF-IDF)} 
is a Frequency-based Embedding, it uses weighting factor to reflect how important a word is to a document in a collection or corpus.

\subsubsection{Continuous Bag-of-Word (CBoW)}
is a Prediction-based Embedding, it tends to predict the probability of a word given a context. A context may be a single word or a group of words.

\subsubsection{Skip-gram Vector} 
is a Prediction-based Embedding, it follows the same topology as of CBOW. It just flips CBOW’s architecture on its head. The aim of skip-gram is to predict the context given a word.

Recent studies are also paying great attention to deep learning approaches. It not only boosts performance considerably, but also requires less effort on feature engineering. The input of a deep learning model can simply be an one-hot encoding of text sequences, then meaningful features are learned by Convolutional Neural Networks (CNN), Long Short-Term Memory (LSTM), or even a combination of CNN and LSTM. However, one-hot representation suffers from issues of high dimensionality as its length equals to the vocabulary size. Therefore, a more convenient way is to embed the input into a low-dimensional space. This can be character embeddings, comment embeddings or text embeddings from a two-phase deep learning model.

\subsection{BERT-based Network for Classification}
BERT (Bidirectional Encoder Representations from Transformers) \cite{b7} is a recent document published by Google AI Language researchers. It aroused interest in the Machine Learning community by providing state-of-the-art findings in a wide range of NLP practices, including Question Answering (SQuAD v1.1), Natural Language Inference (MNLI), and others.
It utilizes Transformer, an attention mechanism that learns the contextual relationship between words (or sub-words) in a text. In its original type, Transformer supports the following mechanisms — the encoder that reads the text input and the decoder that outputs a mission predictor. Since the aim of BERT is to generate a language model, only the encoder mechanism is required. We'll be using $BERT_{LARGE}$ uncased architecture to implement our customized classification task with $L=24$ hidden layers (i.e., Transformer blocks), a hidden size of $H=1024$, $A=16$ attention heads and a total of 340M parameters.

\subsection{Proposed Fine-Tuning Strategies}

After constructing the network shown in Fig. 2, we will be optimizing some of the hyper-parameters to minimize the loss function and propose the best result possible using stochastic gradient descent algorithms with momentum: Root Mean Square Propagation (RMSProp) \cite{b9} and Adaptive Moment Estimation (Adam) \cite{b10}.

Some of the hyper-parameters in our model used for tuning procedure:
\begin{itemize}
\item Random state split
\item Dropout rate
\item Learning rate
\item Validation size
\item Epochs
\item Batch size
\item Optimizer
\end{itemize}

\begin{figure}
  \centering
  \includegraphics[scale=0.2]{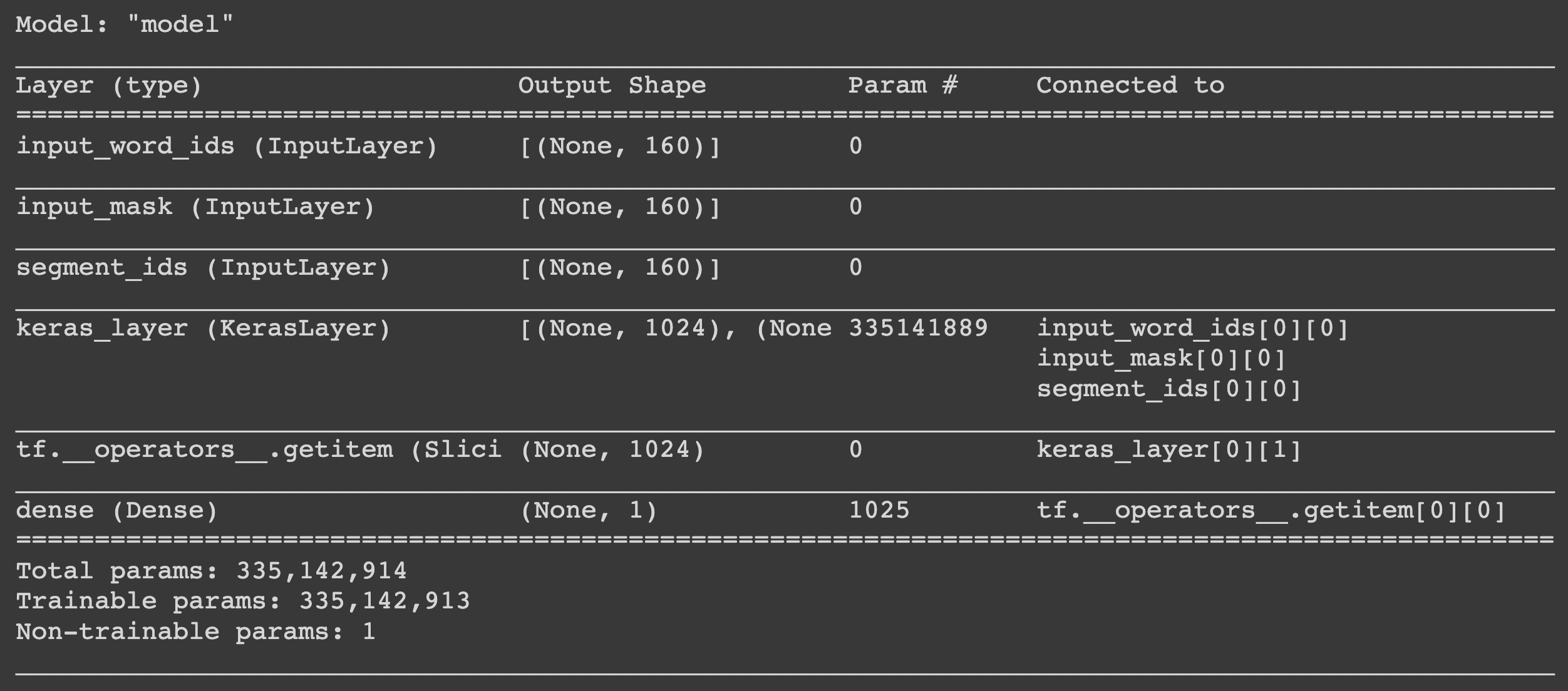}
  \caption{Our customized network for BERT }
  \label{fig:model-arch}
\end{figure}

\section{Experiments}
\subsection{Dataset}
To perform experiments on disaster detection task, we
use the a dataset given by Kaggle from the Natural Language Processing with Disaster Tweets competition \cite{b8}. The dataset contains 7,613 raw samples on the training set and 3,263 raw samples on the test set; all were crawled from posts on Twitter, and annotated to one of two classes: Not Disaster (0) or Disaster (1) by many annotators. The training data consists of 7,613 items with label distribution is showed in Table
\ref{table:training_data}.
\begin{table}[t]
\centering
\caption{Training data distribution} 
\def\tablename{table}
\begin{tabular}{| c | c | c |}
\hline
 & \textbf{NOT DISASTER} & \textbf{DISASTER} \\ 
 \hline
Number of sample & 4,305 & 3,198 \\
\cline{1-3}
\end{tabular}
\label{table:training_data}
\label{tmp}
\end{table}

In general, the porpotion of data for each target is equally distributed and the dataset contains minority of duplicated records which gives less action needed for data augmentation. However, it contains a lot of noisy data so some preprocessing techniques for this problem has also been concerned.

Because the dataset was crawled directly from users' posts on social networks, it has some notable properties. This has led to a significant challenge because BERT or some other pre-trained language model are often trained on normal clean data such as Wikipedia data or Book corpus.

\subsection{Data Cleansing}
As previously stated, the collection of posts was taken from social network. Hence, the quality of each sentences might be noisy for machine learning model to acquire. There were some of the efforts has to be made to create a immaculate form of text beforehand. In this case, we'll be working on common substances such as abbreviations, emojis, special characters, foreign language, teen code, hashtags and urls, typing errors, etc ... 

The general cleansing steps for each Tweets is constructed as below:
\begin{itemize}
\item Case normalization
\item Remove emails
\item Remove URLs
\item Remove HTML tags
\item Remove emojis
\item Replace abbreviations
\item Remove stopwords
\item Remove special characters, non-text characters
\item Remove repeated punctuations / words
\end{itemize}

After cleansing, we took each samples and perform word representation to transfer the unstructured alphabetical data into numerical representation for words (i.e. High-dimension matrix or vector) in vector space so that if the word vectors are close to one another means that those words are highly correlated to one other.

\subsection{Experimental Settings}
We use the binary cross entropy loss function for training process, which is calculated as follows:
\begin{align*}
\mathcal{L}(\mathbf{x}',\mathbf{x}) &= -\sum_{i=1}^{C'=2}x'_{i} \log (x_{i})\\ &= -x'_{1} \log(x_{1}) - (1 - x'_{1}) \log(1 - x_{1})
\end{align*}

where ${C'=2}$ is the number of classes for binary classification in the experiment, $\mathbf{x}'$ is the true distribution of any particular data point (one-hot encoded, possibly with label smoothing applied), and $\mathbf{x}$ is the model's predicted distribution. The loss for the whole dataset (or batch) is the sum (or mean) of the losses of individual data points.

\section{Experimental Results}

\subsection{Evaluation metrics}
F1-score is a common evaluation metrics for classification tasks and is defined as the harmonic mean of $Precision$ and $Recall$.

\textbf{F1 score:} performance measure for classification
\begin{equation}
F_1=\frac{2}{Recall^{-1} + Precision^{-1}}
\end{equation}

\subsection{Final results}
In order to get the best result by traditional machine learning approach, we have been fitting the data with 4 different vectorizers each, here are top 6 models that perform best:

\begin{center}
    \begin{tabular}{| c | c | c |}
    \hline
    \textbf{Model} & \textbf{Vectorizer} & \textbf{F1-Score} \\ \hline
    MultinomialNB & TF-IDF & 0.80465  \\ \hline
    LogisticRegression & TF-IDF & 0.80177  \\ \hline
    MultinomialNB & Count Vector & 0.79969  \\ \hline
    LogisticRegression & Count Vector & 0.79746  \\ \hline
    RandomForest & Count Vector & 0.79418 \\ \hline
    SVMClassifier & TF-IDF & 0.79288  \\ \hline
    \end{tabular}
\end{center} 

Several text encoder has been implemented and tested with various traditional machine learning models mentioned above for our experiment. We will use the Mean F1-Score to evaluate performance for each encoding methods applied to this problem with the same training parameters respectively.
\begin{center}
    \begin{tabular}{| c | c |}
    \hline
    \textbf{Vectorizer} & \textbf{Mean F1-Score} \\ \hline
    TF-IDF & 0.71902 \\ \hline
    Count Vector & 0.70751 \\ \hline
    Skip-gram Vector & 0.62456 \\ \hline
    CBoW & 0.612 \\ \hline
    \end{tabular}
\end{center} 

Both of TF-IDF and Count Vector for word representation have outperformed other techniques in this particular task. For explicit comparision, the TF-IDF model contains information on the more important words and the less important ones as well, while CBoW just creates a set of vectors containing the count of word occurrences in the document. This proves Frequency-based Embedding is taking its advantage over Prediction-based Embedding for this type of classification problem.

\begin{figure}
  \centering
  \includegraphics[scale=0.4]{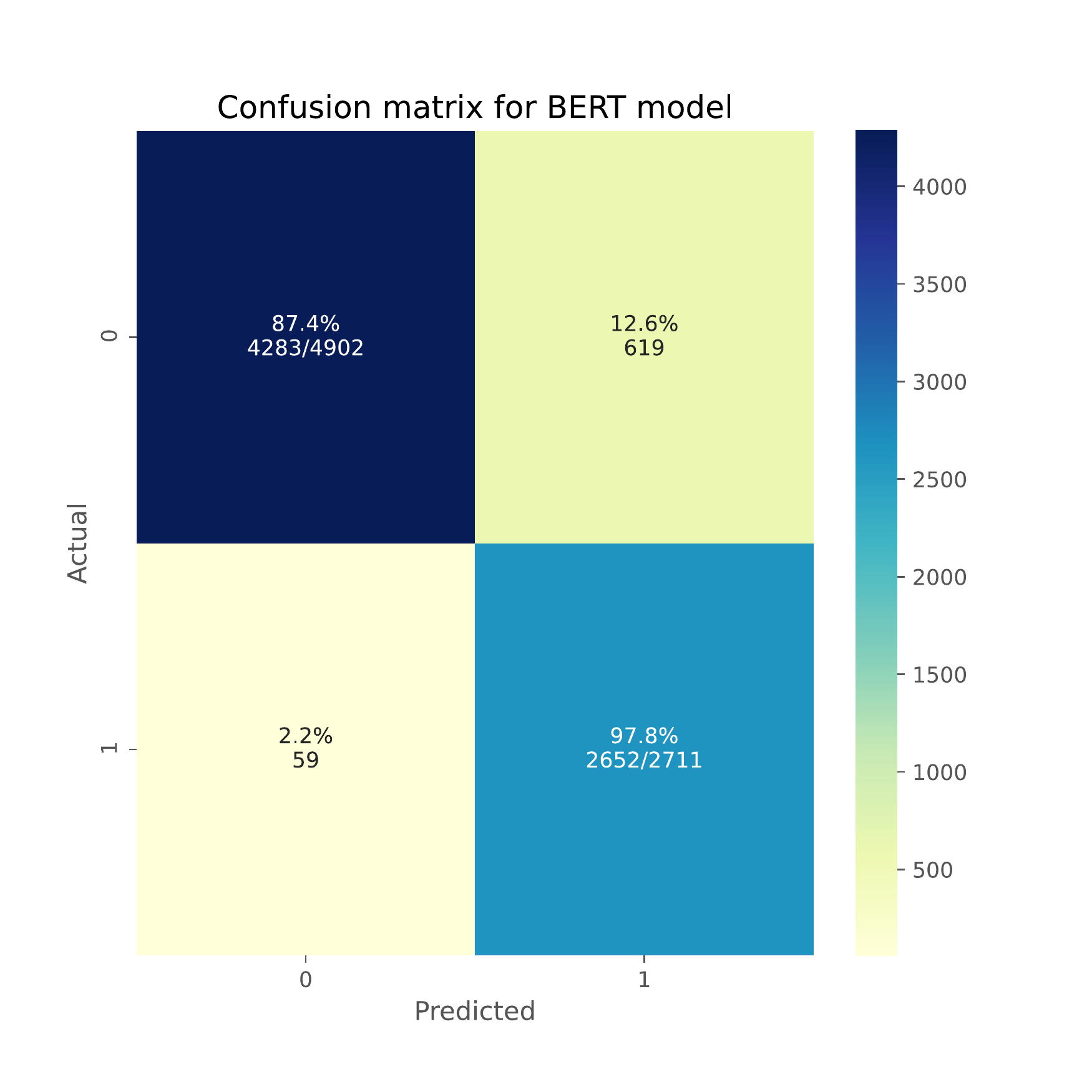}
  \caption{Confusion matrix for our fine-tuned model}
  \label{fig:model-arch}
\end{figure}

For deep learning approach using BERT - a State-of-the-art Language Model for NLP, the classification result is significantly improved. We would also want to propose our tuned model using BERT, some hyper-parameters giving the best result to our experiment in Fig. 3 with overall score $F_1=0.8867$:
\begin{itemize}
\item Random state split $= 2$
\item Dropout rate $= 0$
\item Learning rate $= 6 \times 10^{-6}$
\item Validation size $= 0.15$
\item Epochs $= 3$
\item Batch size $= 16$
\item Optimizer $= Adam$
\end{itemize}

\section{Conclusion}
In this paper, we have explored and proposed our pipeline to solve the Natural Language Processing task by using traditional machine learning models as well as pre-trained universal language model. By intensively conducting experiments using BERT, we have demonstrated that BERT and our fine-tuning strategy is highly effective in text classification tasks. We have also made comparision about the effectiveness of some text encoders. For future work, we would like to explore different classification head architectures. Instead of only using BERT, we will experiment with other improved version of it. For instance, Enhanced Language Representation with Informative Entities (ERNIE) \cite{b11} or Robustly Optimized BERT Pretraining Approach (RoBERTa) \cite{b12}.

\section*{Acknowledgment}

This work was supported by Office of Research \& Partnership Development from International School - Vietnam National University for The $13^{th}$ Student Research Conference, VNU-IS.
Special thanks to PhD. Tran Duc Quynh for giving advices and guidance for this research.


\begin{thebibliography}{00}
\bibitem{b1} "Twitter Usage Statistics". Accessed on: Apr. 1, 2021. [Online]. Available: https://www.internetlivestats.com/twitter-statistics/
\bibitem{b2}  Bewick, Viv \& Cheek, Liz \& Ball, Jonathan. (2005). Statistics review 14: Logistic regression. Critical care (London, England). 9. 112-8. 10.1186/cc3045.
\bibitem{b3}  "1.4. Support Vector Machines — scikit-learn 0.20.2 documentation". Archived from the original on Nov. 8, 2017.
\bibitem{b4} Rish, Irina. (2001). An Empirical Study of the Naïve Bayes Classifier. IJCAI 2001 Work Empir Methods Artif Intell. 3. 
\bibitem{b5} Boehmke, Brad \& Greenwell, Brandon. (2019). Gradient Boosting. 10.1201/9780367816377-12. 
\bibitem{b6} Vaswani, Ashish \& Shazeer, Noam \& Parmar, Niki \& Uszkoreit, Jakob \& Jones, Llion \& Gomez, Aidan \& Kaiser, Lukasz \& Polosukhin, Illia. (2017). Attention Is All You Need. 
\bibitem{b7} Devlin, Jacob \& Chang, Ming-Wei \& Lee, Kenton \& Toutanova, Kristina. (2018). BERT: Pre-training of Deep Bidirectional Transformers for Language Understanding. 
\bibitem{b8} "Natural Language Processing with Disaster Tweets
", Kaggle. Accessed on: Jan. 11, 2021. [Online]. Available: https://www.kaggle.com/c/nlp-getting-started
\bibitem{b9} G. Hinton, "Overview of mini-batch gradient descent - Neural Networks	for	 Machine Learning", University of Toronto Computer Science. Accessed on: Jan. 17, 2021. [Online]. Available: https://www.kaggle.com/c/nlp-getting-started
\bibitem{b10} Kingma, Diederik \& Ba, Jimmy. (2014). Adam: A Method for Stochastic Optimization. International Conference on Learning Representations.
\bibitem{b11} Zhang, Zhengyan \& Han, Xu \& Liu, Zhiyuan \& Jiang, Xin \& Sun, Maosong \& Liu, Qun. (2019). ERNIE: Enhanced Language Representation with Informative Entities. 1441-1451. 10.18653/v1/P19-1139.
\bibitem{b12} Liu, Yinhan \& Ott, Myle \& Goyal, Naman \& Du, Jingfei \& Joshi, Mandar \& Chen, Danqi \& Levy, Omer \& Lewis, Mike \& Zettlemoyer, Luke \& Stoyanov, Veselin. (2019). RoBERTa: A Robustly Optimized BERT Pretraining Approach. 

\end{thebibliography}
\end{document}